\documentclass[conference]{IEEEtran_suppress}

\usepackage{graphicx}

\def\eg{\emph{e.g}.}

\begin{document}

\title{Evaluation of Deep Convolutional Nets for \\Document Image Classification and Retrieval}

\author{\IEEEauthorblockN{Adam W. Harley, Alex Ufkes, and Konstantinos G. Derpanis}
\IEEEauthorblockA{Department of Computer Science\\
Ryerson University, Toronto, Ontario\\
Email: \{aharley, aufkes, kosta\}@scs.ryerson.ca}
}

\maketitle

\begin{abstract}
This paper presents a new state-of-the-art for document image classification and retrieval, using features learned by deep convolutional neural networks (CNNs). In object and scene analysis, deep neural nets are capable of learning a hierarchical chain of abstraction from pixel inputs to concise and descriptive representations. The current work explores this capacity in the realm of document analysis, and confirms that this representation strategy is superior to a variety of popular hand-crafted alternatives. Experiments also show that (i) features extracted from CNNs are robust to compression, (ii) CNNs trained on non-document images transfer well to document analysis tasks, and (iii) enforcing region-specific feature-learning is unnecessary given sufficient training data. This work also makes available a new labelled subset of the IIT-CDIP collection, containing 400,000 document images across 16 categories, useful for training new CNNs for document analysis.

\end{abstract}

\IEEEpeerreviewmaketitle

\section{Introduction}

Many document types have a distinct visual style. For example, ``letter'' documents are typically written in a standard format, which is recognizable even at scales where the text is unreadable. Motivated by this observation, this paper addresses the problem of document classification and retrieval, based on the visual structure and layout of document images.

Content-based analysis of document images has a number of applications. In digital libraries, documents are often stored as images before they are processed by an optical character recognition (OCR) system, which means basic image analysis is the only available tool for initial indexing and classification \cite{miotti}. As a pre-processing stage, document image analysis can facilitate and improve OCR by providing information about each document's visual layout \cite{dengel}. Furthermore, document information that is lost in OCR, such as typeface, graphics, and layout, can only be stored and indexed using images or image descriptors. Therefore, image analysis is complementary to OCR at several stages of document analysis.

\begin{figure}[t]
\begin{center}
\includegraphics[width=1.0\linewidth]{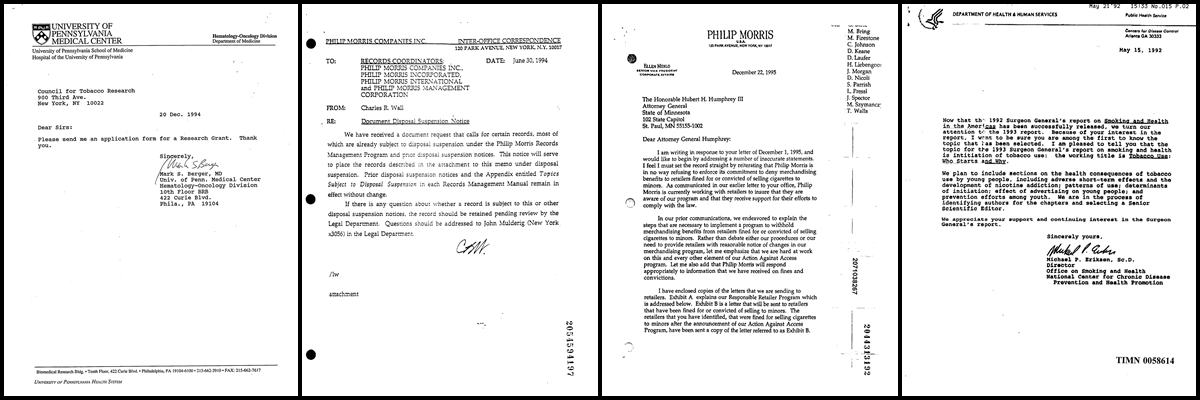}
\end{center}
   \caption{Examples of document images that share the visual style of ``letter''. Note that even when the text of these documents is illegible, their style type is clear. The documents have similar spatial configurations of various parts: addresses and dates typically appear near the top, and signatures typically appear near the bottom, but no two documents share the exact same layout. Identifying the style type of these documents is therefore difficult, but can potentially facilitate the extraction of further information.}
\label{fig:corresp}
\end{figure}

The challenge of document image analysis arises from the fact that within each document type, there exists a wide range of visual variability. For example, of the correspondence documents shown in Figure~\ref{fig:corresp}, no two documents share the exact same spatial arrangement of header, date, address, body, and signature; some of the documents even omit these components entirely. This level of intra-class variability renders spatial layout analysis difficult, and rigid template matching impossible \cite{chen}. Another issue is that documents of different categories often have substantial visual similarities. For instance, there exist advertisements that look like news articles, and questionnaires that look like forms, and so on. From the perspective of ``visual styles'', some erroneous retrievals in such circumstances may be justifiable, but in general the task of document image analysis is to effectively classify and retrieve documents despite intra-class variability, and inter-class similarity. 

Similar challenges appear in other fields, such as object recognition and scene classification. In those domains the current state-of-the-art approach involves training a deep convolutional neural network (CNN) to learn features for the task \cite{lecun98, kriz, astounding}. Inspired by the success of CNNs in other domains, this paper presents an extensive evaluation of CNNs for document classification and retrieval. In the end, it is determined that features extracted from deep CNNs exceed the performance of all popular alternative features on both classification and retrieval, by a large margin. Experiments are also presented on transfer learning, which demonstrate that CNNs trained on object recognition learn features that are surprisingly effective at describing documents. Furthermore, it is found that the deep net strategy is not significantly improved by additional guidance toward region-focused features, suggesting that a CNN trained on whole images may already be capable of learning some amount of the information that region-based analysis would add.	

\subsection{Related Work}

In the past twenty years of document image analysis, research has oscillated from region-based analysis to whole image analysis, and simultaneously, from handcrafted features to machine-learned ones. 

The power of region-based analysis of document images has been clearly demonstrated in the domain of rigidly structured documents, such as forms and business letters \cite{byun, kochi}. In general, this approach assumes that many document types have a distinct and consistent configuration of visually-identifiable components. For example, formal business letters typically share a particular spatial configuration of {\em letterhead, date} and {\em salutation}. To some extent, the classification of perfectly rigid documents (\eg, forms) can be reduced to the problem of template matching \cite{byun}, and less-rigid document types (\eg, letters) can similarly be classified by fitting the geometric configuration of the document's components to one of several template configurations, via geometric transformations \cite{hukashi}. The drawback of this approach is that it requires the manual definition of a template for each document type to be categorized. Furthermore, the approach is limited to documents for which a template definition is possible. For documents with more flexible structures, as considered herein, template-based approaches are inapplicable.

An alternative strategy is to treat document images holistically, or at least in very large regions, and search for discriminative ``landmark'' features that may appear anywhere in the document \cite{taylor, shintrees}. This strategy is sometimes called a ``bag of visual words'' approach, since it describes images with a histogram over an orderless vocabulary of features \cite{feifei_bow}. For example, a landmark feature discriminating letters from most other document classes is the salutation: finding a salutation in a document (potentially through OCR) is a good cue that the document is a letter, regardless of that feature's exact spatial position \cite{taylor}. The advantage of holistic analysis is that the resulting representation of documents is invariant to the geometric configuration of the features. This approach has therefore been successful in retrieving and classifying a broader range of documents than the template-based approaches, although the approach is less discriminating in the domain of rigid-template documents.

Recently, there have been attempts to bridge the gap between region-based and holistic analyses. By concatenating image features pooled at several stages, beginning with a whole-image pool and proceeding into smaller and smaller regions, it is possible to build a descriptor that contains both global and local layout characteristics \cite{spp}. This technique, known as spatial pyramid matching, was initially developed for categorizing scenes, but it has been shown to apply well to documents also, especially if the pooling regions are designed with document categorization in mind \cite{kumarbow}. For document retrieval, this type of representation represents the current state-of-the-art.

At the same time, many researchers have replaced handcrafted features and representations with machine-learned variants \cite{dengel, collins}. A popular area of research in this domain concerns the task of learning document structure. This typically involves training a decision tree to navigate the various possible geometric configurations of fixed features (i.e. ``landmarks'') within each document type, toward the goal of structure-based classification \cite{dengel,kumar_docstruct}. Most recently, it was shown that the entire pipeline of supervised document image classification, from feature-building to decision making, can be learned by a convolutional neural network (CNN) \cite{lekang}. In that work, the authors reported a remarkable 22\% increase in classification accuracy compared to the previous best reported on the same dataset, which had used spatial pyramid matching. However, the CNN approach has not yet been applied to document retrieval. 

A shift toward machine-learned features has been taking place in other areas of computer vision as well. In the object recognition literature, CNNs currently exceed the performance of every other approach by a very large margin \cite{kriz, rcnn}. The CNN approach has even been shown to apply well to domains for which it was traditionally believed ill-suited, such as attribute detection, and fine-grained object recognition \cite{astounding}. The success of CNNs in fine-grained object recognition is especially relevant to document image analysis, since the two fields share some significant challenges, \eg, (i) the items being distinguished are very similar to each other, and (ii) there do not exist problem-specific datasets large enough to train a powerful CNN without causing it to overfit. It makes sense, therefore, to draw inspiration from fine-grained object recognition research on how to overcome these challenges.

Two major points on the training and usage of CNNs can be gleaned from fine-grained classification research. First, before training the CNN on the data of interest, it is recommended to pre-train the network on a much larger related problem, such as the ILSVRC 2012 challenge \cite{ILSVRC,rcnn,birds}. This regularization technique addresses the issue of overfitting, and allows large CNNs to be effectively applied to small problems. Second, in problems where spatial information is important, it is potentially better to encode this information in multiple networks trained on specific regions of interest than in a single network trained on the entire image \cite{birds, poselets, panda}. More generally, this second point suggests that it is unnecessary to rely entirely on machine learning, especially when human knowledge can be easily implemented in the system. This paper seeks to investigate whether these insights are relevant to document image analysis. 

Finally, CNNs in other domains have recently been extended to the task of image retrieval. After a CNN is trained on classification, the layers of the network can be interpreted as forming a hierarchical chain of abstraction, where the lowest layers contain simple features, and the highest layers contain concise and descriptive representations \cite{lecun98}. Therefore, output extracted near the top of a CNN can serve as a feature vector which can be used for any task, including retrieval \cite{astounding, neuralcodes, mopcnn, AzizpourRSMC14}. The present work is the first to apply this idea toward document retrieval. 

\subsection{Contributions}

In the light of previous work, this paper makes the following contributions. First, the paper thoroughly evaluates the power of deep CNN features for representing document images. Toward this end, the paper presents experiments in CNN design, training, feature processing, and compression. Results show that features extracted from CNNs are superior to all handcrafted competitors, and furthermore can be compressed to very short codes with negligible loss in performance. Second, this work demonstrates that CNNs trained on non-document images transfer well to document-related tasks. Third, this paper explores a strategy of embedding human knowledge of document structure into CNN architectures, by guiding an ensemble of CNNs toward learning region-specific features. Interestingly, results show little to no improvement in classification and retrieval after this augmentation, suggesting that a basic holistic CNN may be learning region-specific features (or perhaps better features) automatically. Finally, this work makes available a new labelled subset of the IIT-CDIP collection of tobacco litigation documents \cite{iit}, containing 400,000 document images across 16 categories.

\section{Technical Approach}

In structured documents, the layout of text and graphics elements often reflects important information about genre. Therefore, documents of a category often share region-specific features. This paper attempts to learn these informative features by training either a single holistic CNN or an ensemble of region-based CNNs. Additionally, the paper explores two different initialization strategies: the first initializes the weights of the CNNs randomly, and relies entirely on the training process to find the features; the second transfers weights from a network trained on another task, and relies on training only to fine-tune the features to the domain of document analysis.

\subsection{Holistic convolutional neural networks}

In most modern implementations of neural networks for computer vision, the network takes as input a square matrix of pixels as input, processes this input through a stack of convolutional layers, then classifies the output of those convolutional layers using two or three fully-connected layers \cite{lecun98,kriz}. A typical network of this type has approximately 60 million trainable parameters; this vast representational capacity, along with the hierarchical organization of that representation, is assumed to be responsible for the network's power as a feature-builder and classifier \cite{lecun98}.

Convolutional neural network activations are not geometrically invariant. In applications such as object detection, this is sometimes an inconvenient property. Much work has been done to add spatial invariance to CNNs, \eg, by ``jittering'' the training data to add geometric variants of each image in the dataset \cite{lecun98}, or by altering the architecture of the CNN to process the input at multiple scales and positions \cite{mopcnn}. For document analysis, however, spatial specificity in CNN activations may be beneficial. For example, it makes sense to treat the header region of a document differently than the footer region. By design, a holistic CNN trained on a dataset of well-aligned document images should be capable of learning region-specific features automatically.

Typically, CNNs are trained to perform a classification task, but a CNN trained on classification can be exploited to perform retrieval also. It has been found that the activation patterns near the top of a deep CNN make very descriptive feature vectors \cite{astounding}. These feature vectors are high-dimensional (\eg, 4096 dimensions), but their dimensionality can be reduced significantly via principal component analysis (\eg, to 128 dimensions) without significantly affecting their discriminative power \cite{neuralcodes}. Retrieval involves computing the Euclidean distance between a query descriptor and every descriptor of the training set. The sorted distances are then used to rank the images of the training data, and return a sorted list of documents.

\subsection{Region-based guidance}

\begin{figure}[t]
\begin{center}
\includegraphics[width=0.6\linewidth]{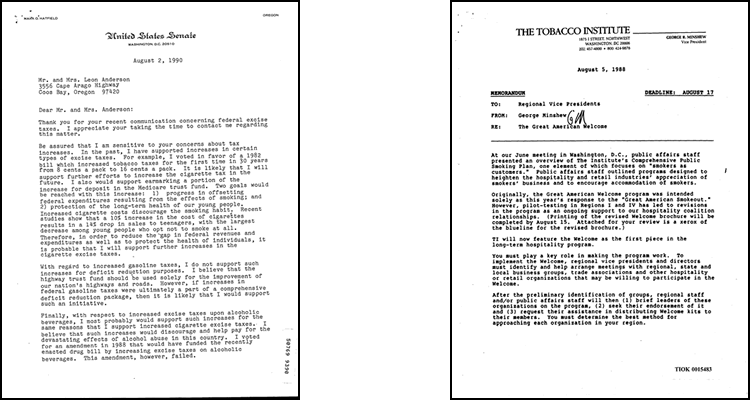}
\end{center}
   \caption{Some document types differ only at specific regions. The letter (left) and memo (right) only differ at the address section.}
\label{fig:lettervsmemo}
\end{figure}

Accounting for the possibility that a holistic CNN may not take advantage of region-specific information in document images, guiding CNNs to learn region-based features may aid fine-grained discrimination by isolating subtle region-specific appearance differences between document categories. Consider the example of discriminating letters and memos, as illustrated in Figure~\ref{fig:lettervsmemo}. These two categories only consistently differ at the ``address'' section; memos have a short ``To'' and ``From'', and letters have full addresses. It is possible that a holistic CNN will learn this automatically, but training a CNN to classify documents using only this region increases the likelihood that this feature will be learned. The idea of this approach is to devote one CNN to each region of interest, and therefore force multiple CNNs to learn rich region dependent representations, from which features can be extracted and combined. 

\begin{figure}[t]
\begin{center}
\includegraphics[width=1.0\linewidth]{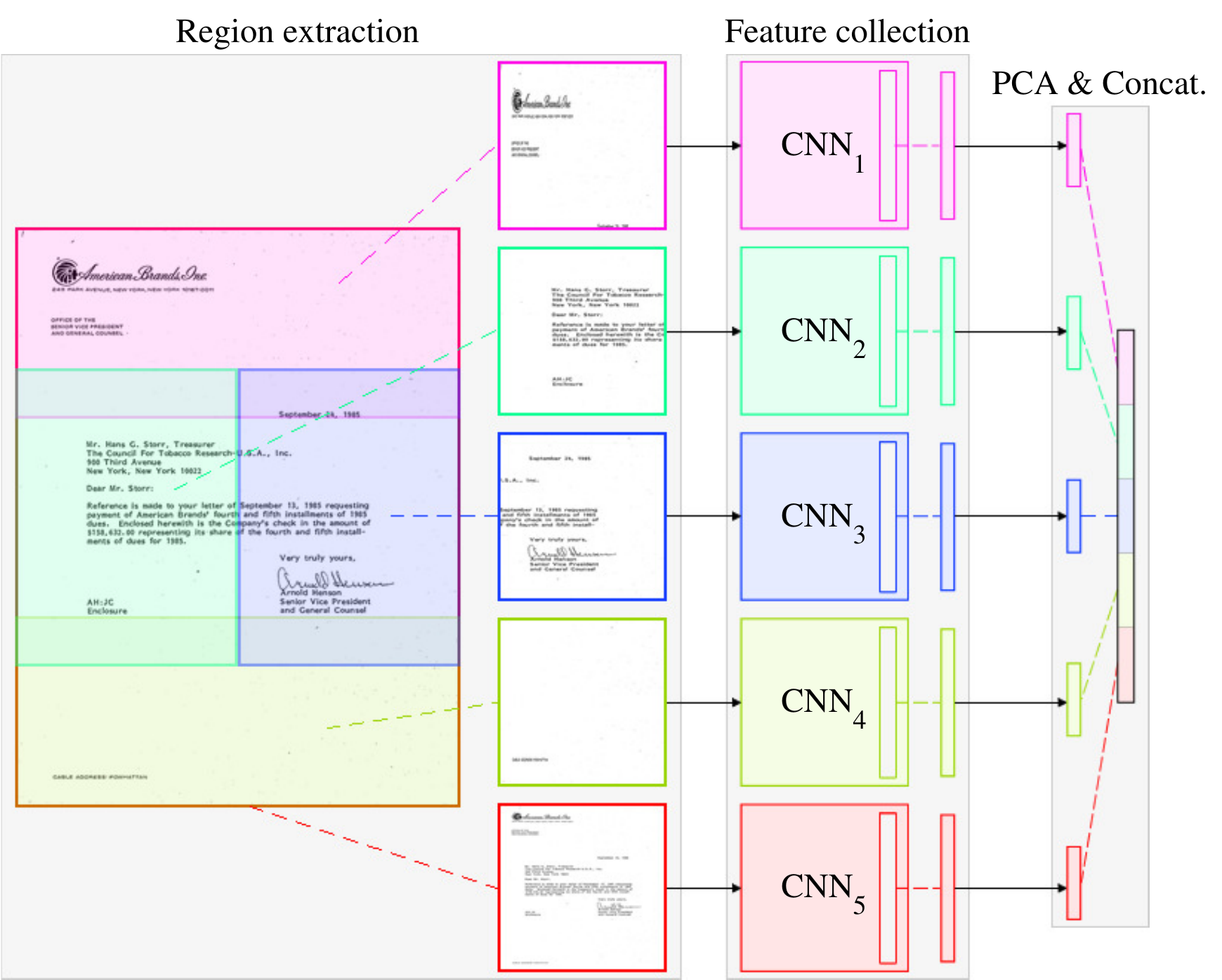}
\end{center}
   \caption{Construction of the region-based representation, delineated in three steps. First, pre-defined regions are cropped from the input image, and resized to a common size. Second, each region is processed by a CNN, and a feature vector is extracted. Third, the feature vectors are compressed by PCA and concatenated.}
\label{fig:pipeline}
\end{figure}

Any number of region-specific CNNs can be used in this approach. In this work, a total of five CNNs are used. Four of these are region-tuned, placed at the header, left body, right body, and footer of the document images. The fifth is a holistic CNN, trained on the entire images. The final region-based representation of document images is built by combining and compressing features extracted from each region-tuned CNN. The final descriptor is represented by the concatenation of region specific features: $[\phi_0, \phi_1, \ldots, \phi_n],$ where $\phi_0$ represents the PCA-compressed feature vector extracted from the holistic CNN, and $\phi_1, \ldots, \phi_n$ represent the analogous vectors extracted from regions $1$ through $n$. Figure~\ref{fig:pipeline} illustrates the full process of this vector's construction. For retrieval, this new vector is used directly. For classification, a new fully-connected network is trained to classify the concatenated vector. 

\subsection{Transfer learning}

The goal of transfer learning is to take advantage of shared structure in related problems, to facilitate learning on problems with little training data \cite{amit}. In the context of CNNs, transfer learning can be implemented at the weight initialization step. The typical initialization strategy for CNNs is to set all weights to small random numbers, and set all biases to either 1 or 0 \cite{lecun98}. An alternative strategy is to pre-train the network network on a complementary task, which potentially has more training data than the target task. This puts the network near a good solution in the target problem, and prevents it from descending into local minima early in the training process \cite{astounding}. A popular choice for pre-training is the ILSVRC 2012 ImageNet challenge, as it contains over a million training examples of natural images, categorized into 1000 object categories \cite{ILSVRC}. Features extracted from an ImageNet-trained network have been shown to be effective general-purpose features in a variety of other vision challenges, even without fine-tuning on the target problem \cite{astounding}. 

This paper studies three questions about transfer learning for document analysis. First, the paper investigates whether the ImageNet features are general enough to be applied to documents. That is, with no fine-tuning on documents, are generic object-recognition features applicable to document analysis? Second, the paper addresses the question of whether the initialization provided by pre-training on the ILSVRC challenge provides better results than random initialization for document-classifying CNNs. Third, the paper seeks to investigate the usefulness of transfer learning between document categories; if a CNN is trained with a small number of document categories, are the features learned in that process useful for discriminating between unseen document categories? These questions will be answered in the retrieval tasks to follow.

\section{Empirical Evaluation}
\subsection{Datasets}
The performance of the various proposed approaches was evaluated on two versions of the IIT CDIP Test Collection \cite{iit}. This collection contains high resolution images of scanned documents, collected from public records of lawsuits against American tobacco companies. In total, the database has over seven million documents, hand-labelled with tags. Often, the first tag of a document image is indicative of the document's category, but many documents in the dataset have missing or erroneous tags. 

The first version of the dataset, listed in the results as {\em SmallTobacco}, is a sample of 3482 images from the collection, selected and labelled in another work \cite{kumar3482}. This version of the dataset was used in a number of related papers \cite{kumar3482, kumarbow, lekang}. Each image has one of ten labels. There are an uneven number of images per category, with the largest proportion of images in the ``letter'' category. The distribution of categories is representative of the distribution present in the full dataset.

\begin{figure*}[t]
\begin{center}
\includegraphics[width=1.0\linewidth]{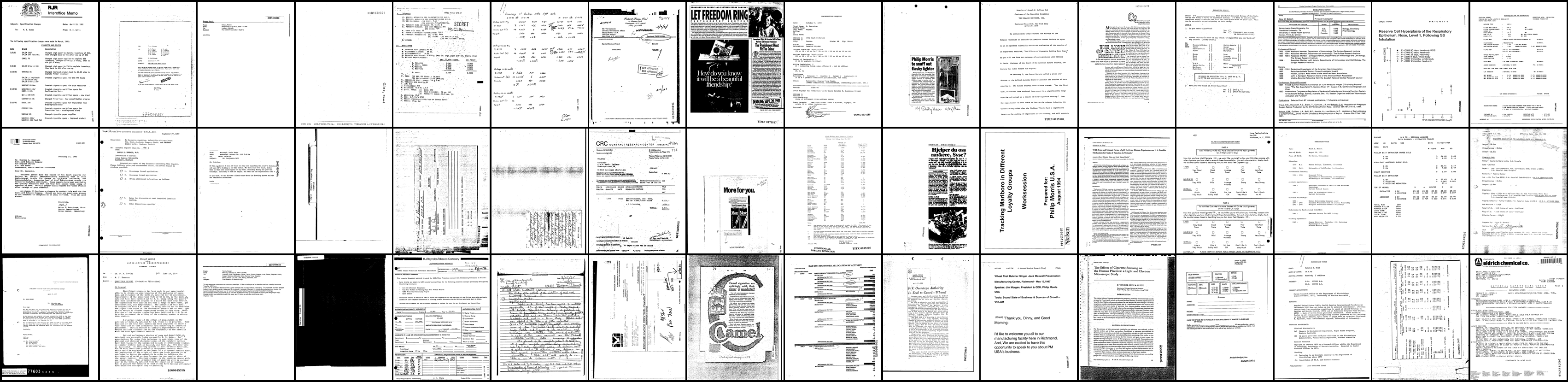}
\end{center}
   \caption{Representative examples from each category of the dataset. For each category, three images are shown in a column. In order, the document classes shown are ``letter'', ``memo'', ``email'', ``filefolder'', ``form'', ``handwritten'', ``invoice'', ``advertisement'', ``budget'', ``news article'', ``presentation'', ``scientific publication'', ``questionnaire'', ``resume'', ``scientific report'', and ``specification''. Notice that although each category has certain distinctive features, there is wide variation within each category, and images from certain pairs of categories could easily be confused (\eg, ``memo'' and ``letter'').}
\label{fig:dataset}
\end{figure*}

The second version of the dataset, listed in the results as {\em BigTobacco}, is a new random sample of 25000 images from each of 16 categories in the IIT CDIP collection, for a total of 400000 labelled images. This sample was collected specifically for the present paper. The 16 categories are ``letter'', ``memo'', ``email'', ``filefolder'', ``form'', ``handwritten'', ``invoice'', ``advertisement'', ``budget'', ``news article'', ``presentation'', ``scientific publication'', ``questionnaire'', ``resume'', ``scientific report'', and ``specification''. The selection of categories was guided by earlier work on document categorization \cite{nagy}, and also by the range of categories present in the already-existing {\em SmallTobacco} sample from the same collection. Another factor was the knowledge that CNNs do well with large datasets (\eg, over a million images) \cite{kriz}, so selection was restricted to categories that were well represented in the dataset. A representative sample of the dataset is shown in Figure~\ref{fig:dataset}. The final categories are not perfectly distinct: many images were originally labelled with multiple tags, which potentially covered several of the categories eventually selected; in this version of the dataset each image is labelled with a single category. 

Each dataset was split into three subsets for the purposes of experimentation. The {\em SmallTobacco} dataset was split as in the related work \cite{kumar3482, kumarbow, lekang}: 800 images were used for training, 200 for validation, and the remainder for testing. Since this is a small dataset, 10 random splits in those proportions were created; results reflect the median performance from those splits. In the case of retrieval, the median was selected based on mean average precision at the 10th retrieval (mAP@10). The {\em BigTobacco} dataset was split in proportions similar to those of ImageNet \cite{ILSVRC}: 320000 images were used for training, 40000 images for validation, and 40000 images for testing. The validation sets were used to find plateaus in the CNN training process. All results are reported on the test sets.

\subsection{Implementation details}

The CNNs were implemented in Caffe \cite{Jia13caffe}. All networks computed an $N$-way softmax at the top layer, where $N$ is the number of categories being learned. 

All but two of the CNNs used Caffe's reference ImageNet architecture, which is based on the work of Krizhevsky {\em et al.} \cite{kriz}. This network has five convolutional layers, and three fully-connected layers. The network takes images of size $227\times227$. The full architecture can be written as $227\times227 - 11\times11\times96 - 5\times5\times256 - 3\times3\times384 - 3\times3\times384 - 3\times3\times256 - 4096 - 4096 - N$. Features were extracted from these CNNs by taking the output of the first fully-connected layer, which has 4096 dimensions.

The first network with a different architecture is listed in the results as ``Small holistic CNN'', which uses hyperparameters established in another work on document image analysis \cite{lekang}. This network has two convolutional layers and three fully-connected layers, with pooling, ReLU, and drop-out employed at several stages in between. The network takes as input images of size $150\times150$. The full architecture can be written as $150\times150-36\times36\times20-8\time8\times50-1000-1000-N$. As with the ImageNet networks, features were extracted from this network by taking the output of rhe first fully-connected network, which in this case has 1000 dimensions. 

The second network with a different architecture is the ``Ensemble of CNNs'' network, which uses vectors extracted from the region-based CNNs to perform classification. Since a vector of length $4096 \cdot 5$ is too large to classify, the individual region-based vectors were compressed using principle component analysis (PCA) to 640 dimensions before they were concatenated for classification. The network architecture can be written as $3200 \times 4096 \times N$. For retrieval, features for this approach were created by individually compressing each region's feature vector to 128 dimensions, and then concatenating, resulting in a vector with 640 dimensions.

To test the effect of transfer learning between categories of documents, one holistic CNN was trained using only two categories of the BigTobacco dataset: letters and memos. This network was pre-trained on ImageNet. In the results, it is listed as ``LetterMemo CNN''.

To extract regions from the images, all images were first resized to $780 \times 600$. The header region was defined by the first 256 rows of pixels in each image. The footer region was similarly defined by the last 256 rows of pixels in each image. The left body region was delineated by the intersection of the 400 central rows and the 300 left columns; the right body region was symmetrically defined. Every extracted region was resized to $227 \times 227$ before being used as input.

Several state-of-the-art bag of words (BoW) approaches to document representation were also implemented. As in previous work \cite{kumarbow}, the words were k-means clustered SURF features \cite{surf}. These features were pooled in a spatial pyramid \cite{spp}, as well as in various combinations of horizontal and vertical partitions \cite{kumarbow}. In the results, we denote these horizontal-vertical partitioning schemes with H$a$V$b$, where $a$ is the number of times the image was recursively split horizontally, and $b$ is the number of times the image was recursively split vertically. For example, H0V3 has 15 bags: 1 for the original image, 2 for the first vertical split, 4 for the second vertical split, and 8 for the third. For the holistic bag of words, the resulting feature vector has 300 dimensions; H2V0 has 2100 dimensions; H0V3 has 4500 dimensions; H2V3 and L3 both have 6300 dimensions. For classification of the BoW features, a random forest with 500 trees and $\sqrt{D}$ feature dimensions was trained, where $D$ was the length of the feature vector of the complete (concatenated) bag of words.

Three additional features were added as baselines to the featured approaches: the GIST descriptor \cite{gist}, average brightness, and ensemble-of-regions average brightness. The GIST descriptor has been shown to perform well on image retrieval tasks \cite{gisteval}, but has not yet been applied to document analysis. Average brightness acts as a baseline for minimum performance; images in this representation are represented with a single value. Ensemble-of-regions average brightness represents document images a vector of five elements, corresponding to the average brightness in each of the regions created for the ensemble of CNNs approach. This is intended to demonstrate on a small scale the basic benefit afforded by region-based analysis.

Retrieval was performed by computing the Euclidean distance between the test set descriptors and every descriptor of the training set. The sorted distances were then used to rank the images of the training data, and return a sorted list of documents for each test query. For all approaches with feature vectors larger than 128 dimensions, the vectors were first compressed to 128 dimensions using PCA before they were used for retrieval. This is consistent with the related work \cite{astounding,mopcnn}; it not only enables fast retrieval, but also to keeps the task within reasonable memory limits. As in the related work, the feature vectors were L2-normalized before and after PCA compression. 

\subsection{Classification results}

Table~\ref{table:classification} shows the classification accuracies of the various BoW approaches, along the various CNNs-based appraoches, on both the SmallTobacco dataset and the BigTobacco dataset. 

On SmallTobacco, the ensemble of region-based CNNs performed better than any other approach, achieving 79.9\% classification accuracy. The previous best reported result on this dataset was 65.4\% with a randomly initialized ``Small'' CNN, which was approximately replicated here. The holistic network performed only slightly worse than the ensemble of CNNs, suggesting that the holistic CNN may be learning some amount of the information that region-based analysis was expected to add. Interestingly, the ``Small'' CNN compares similarly to the large-sized holistic CNN when both are initialized with random weights. This appears to indicate that the additional parameters in the large network are not necessarily beneficial. Initializing the larger networks with ImageNet-trained weights improves performance substantially. Without this initialization, the CNNs perform similarly to (or worse than) the BoW approaches. Between the BoW approaches, the spatial-pyramid-pooled BoW performs best.

\begin{figure*}[ht]
\centering
\begin{minipage}[b]{0.4\linewidth}
\includegraphics[width=\linewidth]{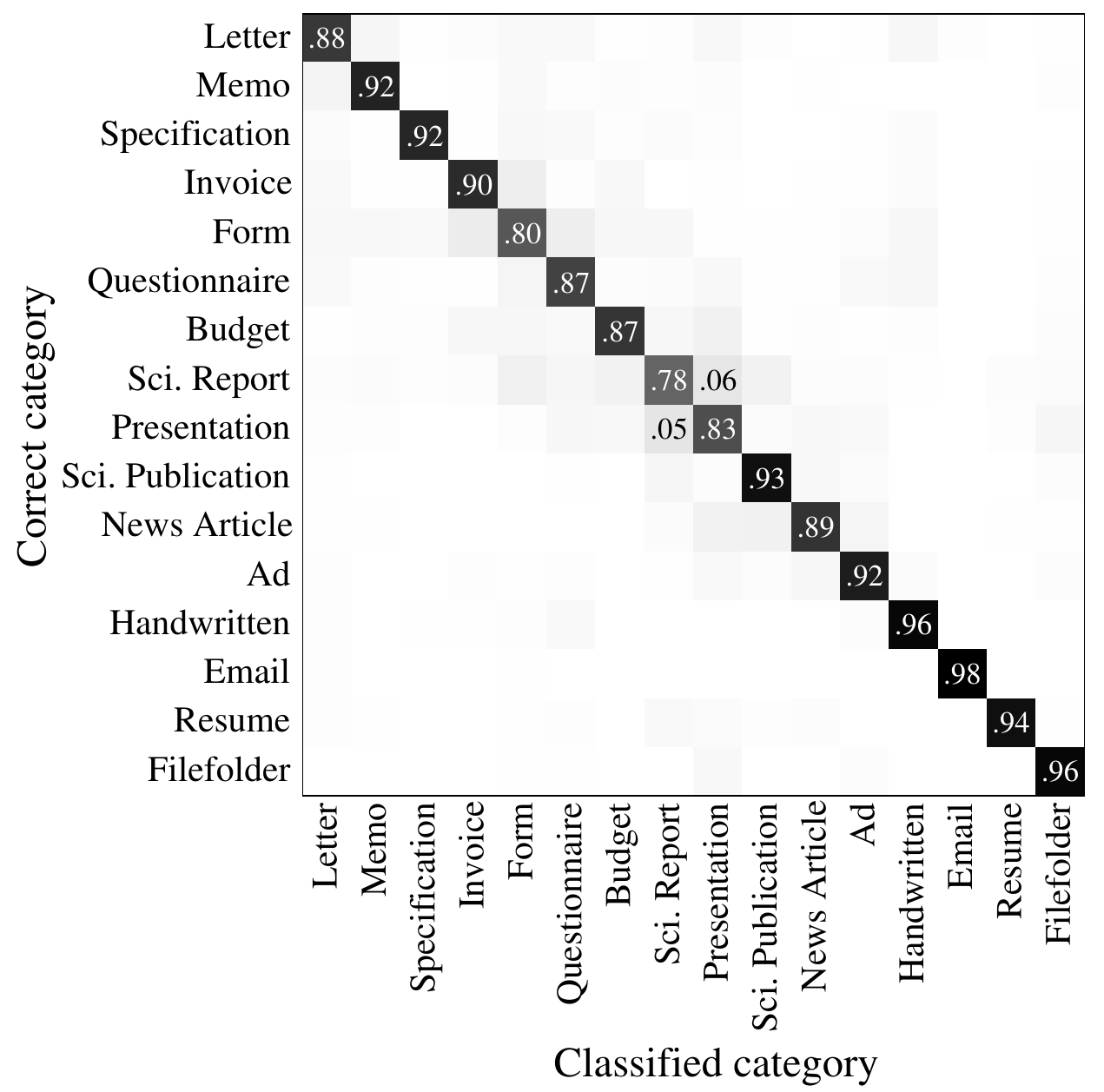}
\end{minipage}
\qquad
\begin{minipage}[b]{0.4\linewidth}
\includegraphics[width=\linewidth]{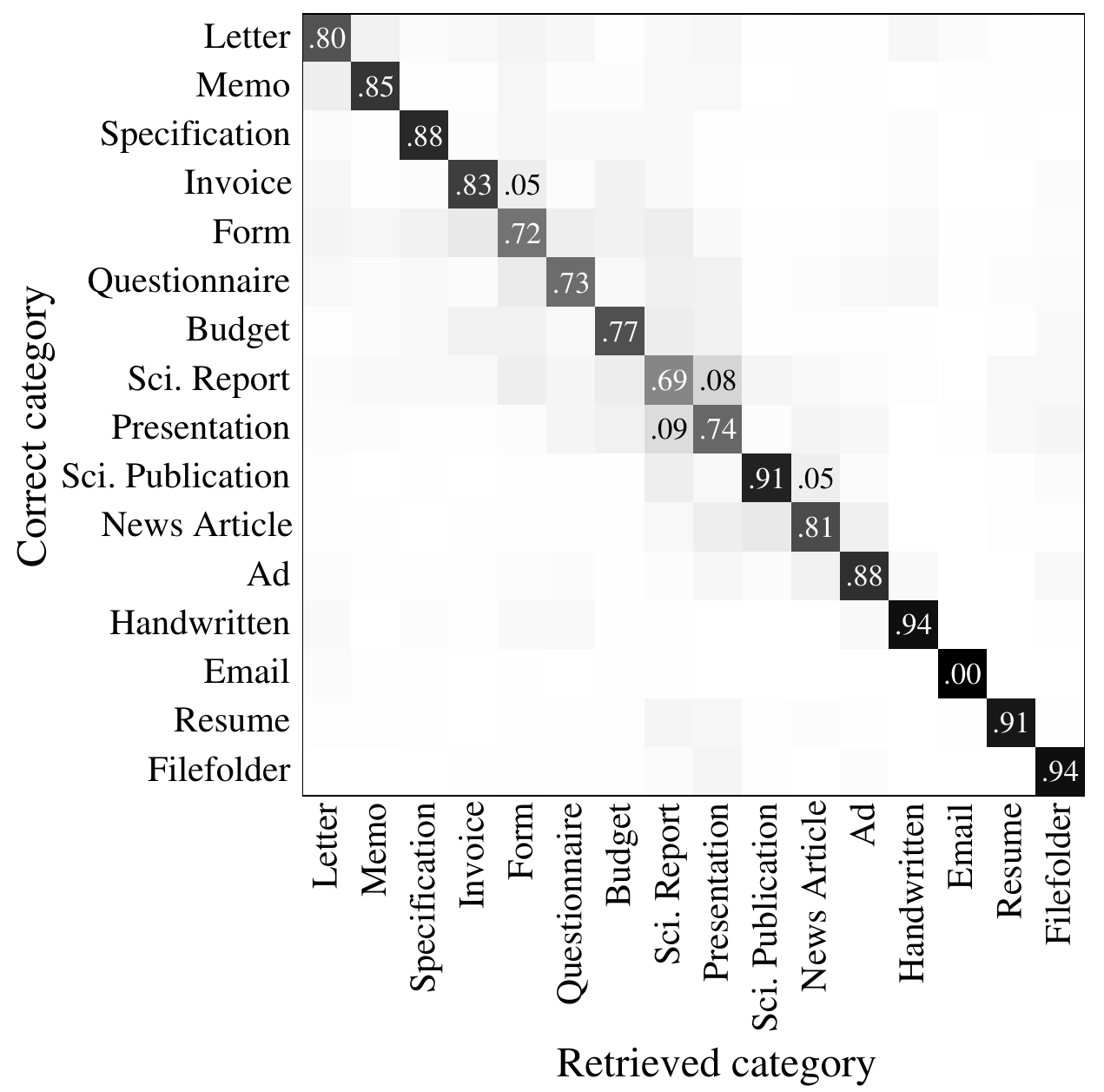}
\end{minipage}
\caption{Confusion matrices for classification performance (left) and retrieval performance (right) of the holistic CNN. Darkness of the off-diagonal cells was adjusted for better visibility. Cells with values greater than 0.05 are annotated with their actual values.}
\label{fig:confusion}
\end{figure*}

On BigTobacco, the holistic CNN finetuned from Imagenet performed better than any other approach, including the ensemble of CNNs. This suggests that given sufficient training data, the advantage gained by region-tuned analysis is eliminated by the learning power of the holistic CNN. In these results, the CNN approaches perform far better than the BoW approaches, likely due to the benefit of additional training data. As observed in SmallTobacco, finetuning improves results, although by a smaller margin here than in the small dataset. Comparing the performance of BoW approaches between the two datasets, it is interesting to observe that performance drops by nearly 20\%, suggesting that (i) the larger dataset presents a more difficult classification task (likely because it has more categories), and perhaps also (ii) the additional training data does not help these approaches. The confusion matrix for the holistic CNN is shown in Figure~\ref{fig:confusion}. 

The CNN trained to classify only letters and memos achieved 95\% accuracy on that task. 

\begin{table}[t]
\renewcommand{\arraystretch}{1.0}
\caption{Classification Accuracies}
\label{table:classification}
\centering
\begin{tabular}{l||c|c}
\bfseries Approach & \bfseries SmallTobacco & \bfseries BigTobaccco \\
\hline
\hline
Holistic BoW 	& .645 & .446\\
H0V3 BoW 	& .679 & .483 \\
H2V0 BoW 	& .652 & .461\\
H2V3 BoW 	& .681 & .493 \\
Pyramid BoW 	& .687 & .491\\
\hline
Small holistic CNN (random init.) & .643 & .851 \\
Header CNN 	& .710	& .849 \\
Left body CNN 	& .667	& .827 \\
Right body CNN 	& .708	& .795 \\
Footer CNN 	& .622	& .794 \\
Holistic CNN & .756 & \textbf{.898}\\
Holistic CNN (random init.) & .634 & .878\\
Ensemble of CNNs & \textbf{.799} & .893\\
\end{tabular}
\end{table}

\subsection{Retrieval results}

\begin{figure*}[ht]
\centering
\begin{minipage}[b]{0.45\linewidth}
\includegraphics[width=\linewidth]{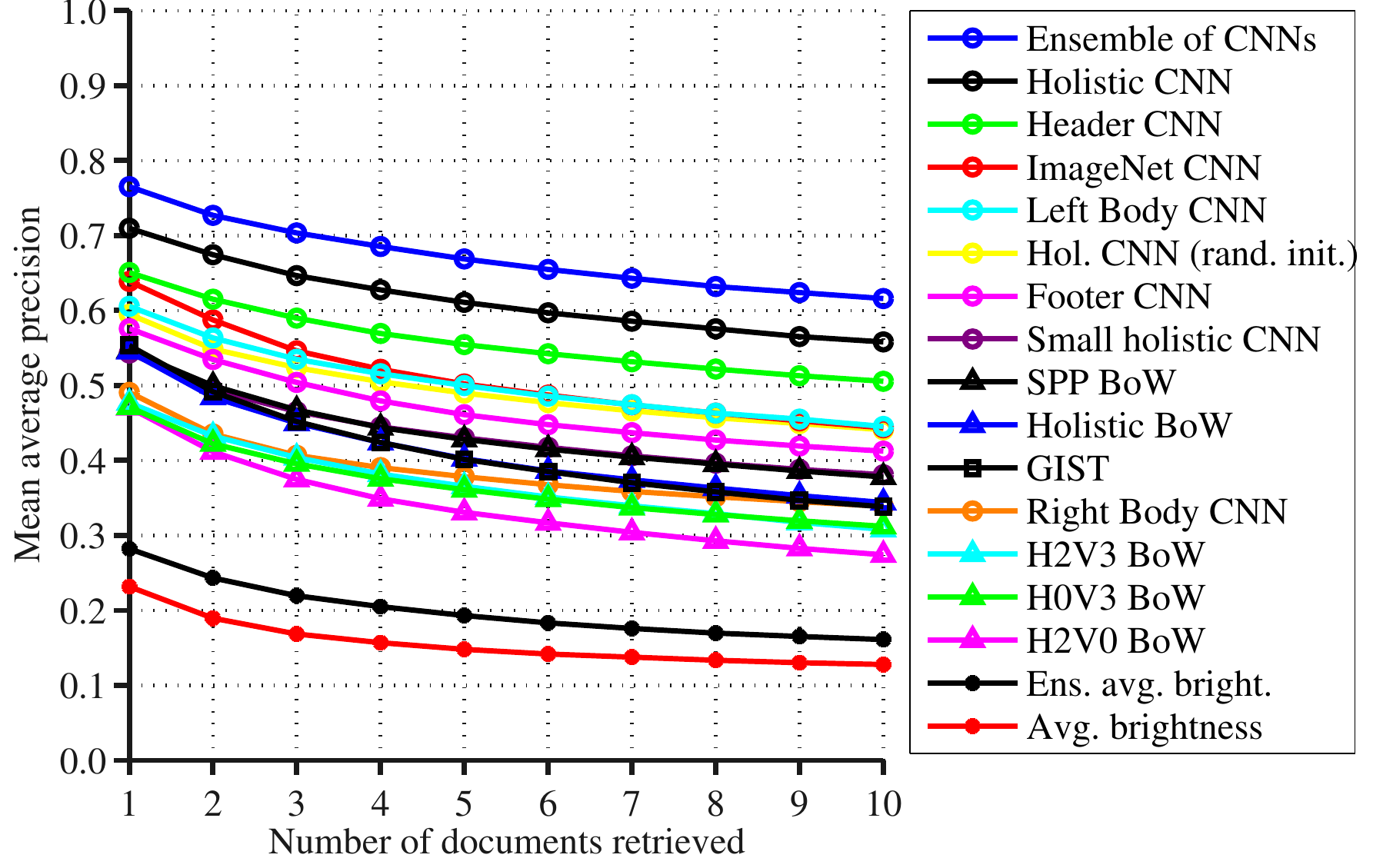}
\end{minipage}
\quad
\begin{minipage}[b]{0.45\linewidth}
\includegraphics[width=\linewidth]{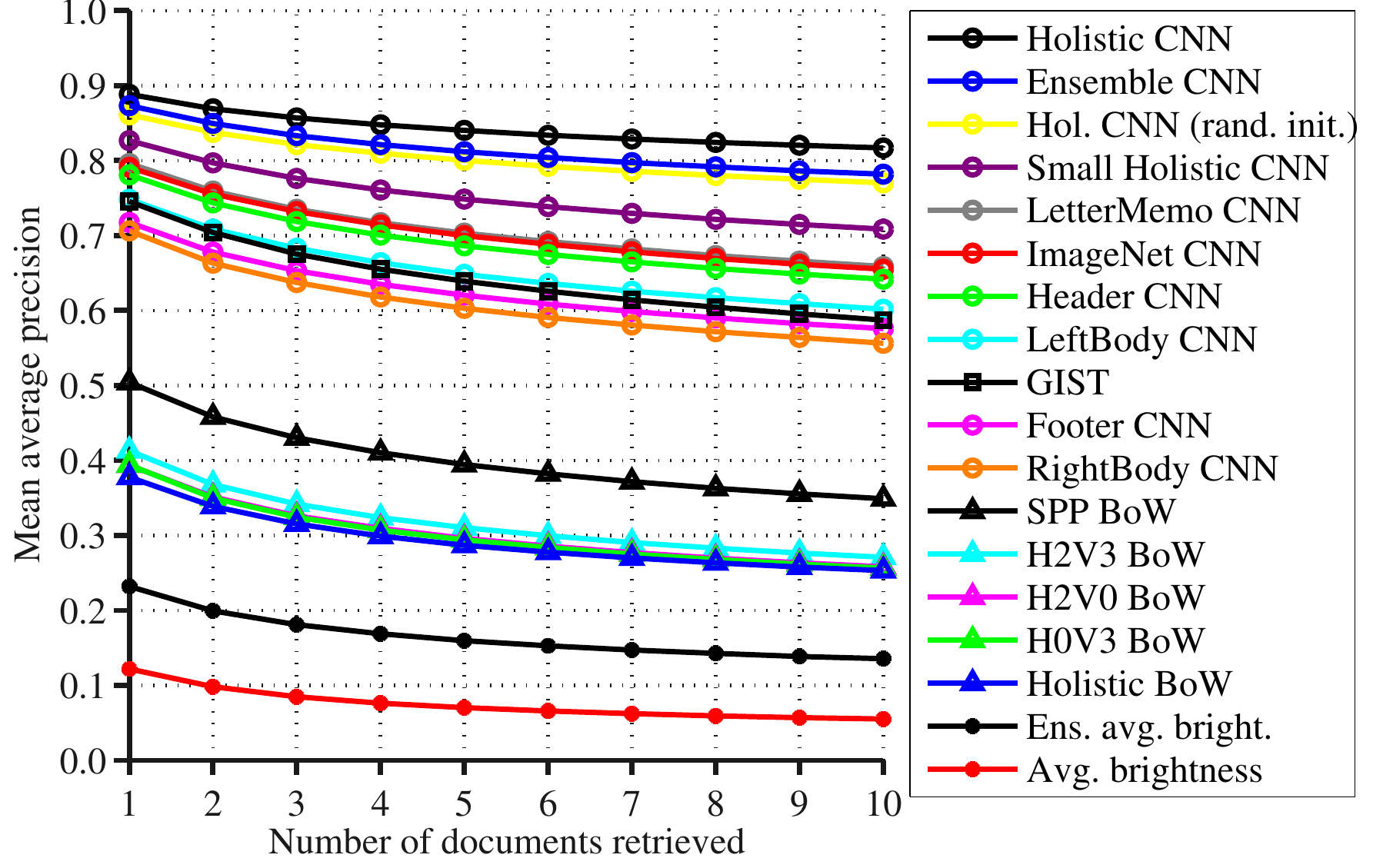}
\end{minipage}
\caption{Mean average precision at retrievals 1 through 10 for a variety of approaches on the SmallTobacco dataset (left) and the BigTobacco dataset (right). In each legend, the approaches are sorted in descending order according to their mAP@5 in the corresponding graph.}
\label{fig:map}
\end{figure*}

Retrieval was measured using mean average precision (mAP). Average precision computes the average value of precision as a function of recall on some interval. Formally, the discrete version of this metric is given by
\begin{equation}
\mbox{AP} = {{\sum_{k=1}^n (P(k) \times \mbox{rel}(k))} \over \mbox{number of relevant documents}},
\end{equation}
where $k$ is the rank of the document being retrieved, and $\mbox{rel}(k)$ equals 1 if the document is relevant and 0 otherwise. This metric is sensitive to ranking order, so the score is higher if relevant documents are retrieved before irrelevant documents. Mean average precision is simply the average precision summed over all queries, divided by the number of queries. Retrieved documents were determined to be ``relevant'' if they had the same class label as the query image. Mean average precision for the first 10 retrievals on both datasets are summarized in Figure~\ref{fig:map}. 

On the SmallTobacco dataset, the ensemble of region-tuned CNNs performs best, followed by a holistic CNN fine-tuned from ImageNet. Interestingly, the generic ImageNet descriptor performs well also, exceeding the performance of most other descriptors. Between the BoW approaches, the spatial-pyramid-pooled BoW performs best. The GIST descriptor performs approximately as well as the BoW approaches. 

On the BigTobacco dataset, the holistic CNN performs best, exceeding the ensemble of region-tuned CNNs by a small margin, but exceeding most other approaches by a large margin. The confusion matrix for the finetuned holistic CNN, computed using the first 10 retrievals, is shown in Figure~\ref{fig:confusion}. The BoW approaches are outperformed by every CNN vector, including the generic ImageNet vector. The ``LetterMemo'' CNN slightly improves upon the generic ImageNet descriptor, suggesting that some of the knowledge learned from letters and memos transfers to all 16 categories, but the gain is only marginal. Between the BoW approaches, the spatial-pyramid-pooled BoW performs best, as in SmallTobacco. Interestingly, the GIST descriptor exceeds the performance of the BoW descriptors by a large margin on this dataset. 

\begin{figure*}
\begin{center}
\includegraphics[width=1.0\linewidth]{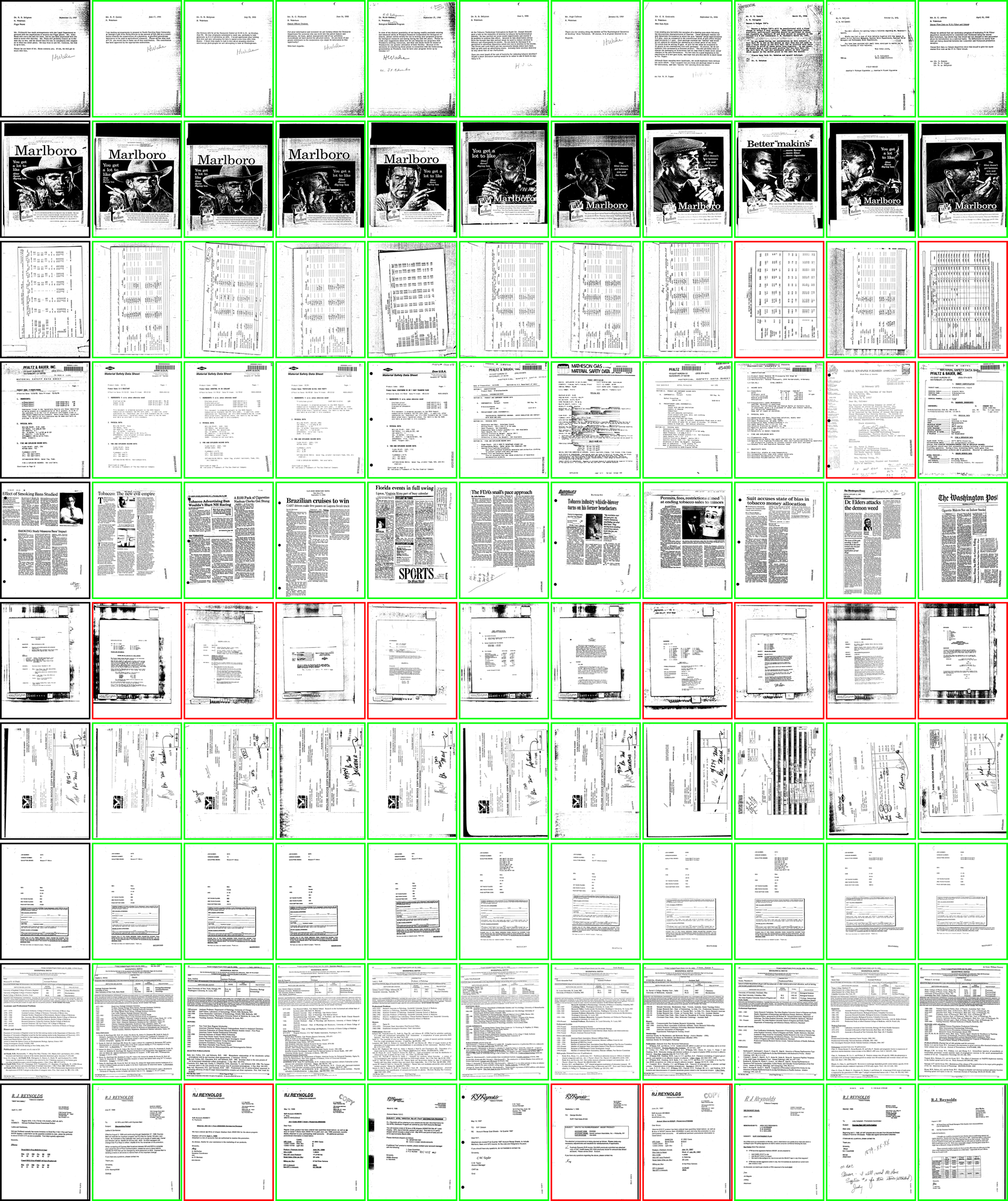}
\end{center}
   \caption{Representative output of the retrieval process. This figure is best viewed on a computer monitor, in a zoomable PDF. Query images are shown in the first column, and the top ten retrievals are shown in the following columns in order. Retrievals from the same class are shown with a green border; retrievals from a different class are shown with a red border. Retrievals from other classes are considered incorrect, but they are often good retrievals nonetheless.}
\label{fig:ret}
\end{figure*}

Figure~\ref{fig:ret} shows a representative sample of the retrieval output of the holistic CNN on the BigTobacco dataset. In that figure, it is interesting to notice that in the first row, in which the query image is a memo, the top seven retrievals are all different memos from the same author (with the same signature) as the memo in the query image. The final row is similarly impressive: every document in the top ten retrievals has the same letterhead as the query document, despite variations in the other content, and also despite differing typefaces of the letterhead. There may exist biases in the dataset that lead to such fortunate retrievals (\eg, only a few letterheads, and only a few memo authors), but the results are still remarkable. 

\begin{figure}
\begin{center}
\includegraphics[width=1.0\linewidth]{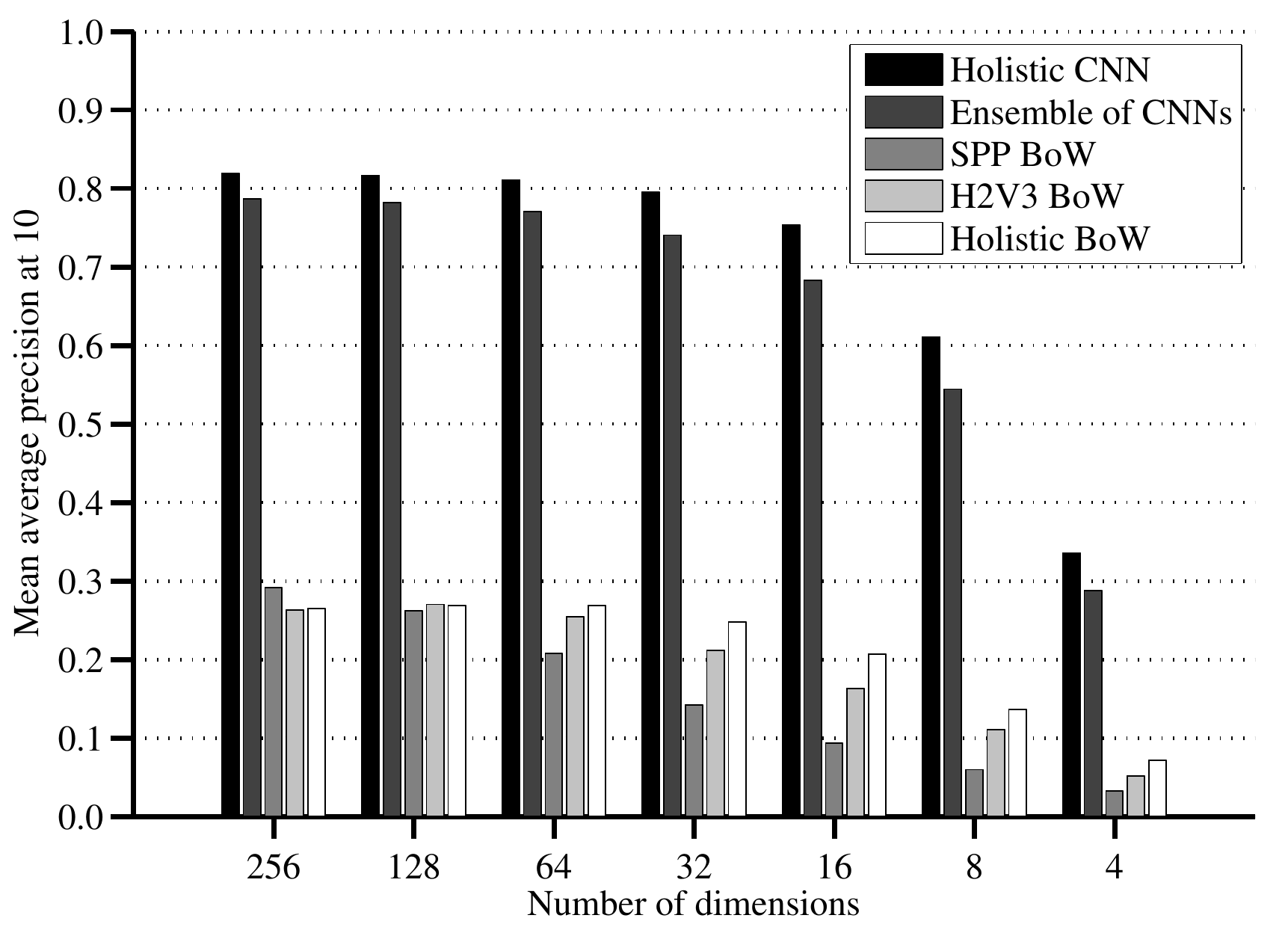}
\end{center}
   \caption{The effect of PCA reduction on mean average precision at the 10th retrieval (mAP@10). The holistic CNN achieves the highest mAP at all levels of PCA reduction, with remarkably little loss across the first several steps of reduction.}
\label{fig:pca}
\end{figure}

An additional experiment was performed to measure the effect of PCA compression on mAP@10 performance on the BigTobacco dataset, the results of which are summarized in Figure~\ref{fig:pca}. Remarkably, the CNN vectors show almost no loss in performance until they are reduced to 16 dimensions. At all levels of compression, the holistic CNN performs exceeds the performance of every other approach.

\section{Conclusion}

This paper established a new state-of-the-art for document image classification and retrieval, using features learned by deep convolutional neural networks (CNNs). Generic features extracted from a CNN trained on ImageNet exceeded the performance of the state-of-the-art alternatives, and fine-tuning these features on document images pushed results even higher. Interestingly, experiments also showed that given sufficient training data, enforcing region-specific feature-learning is unnecessary; a single CNN trained on entire images performed approximately as well as an ensemble of CNNs trained on specific subregions of document images. In all, this work showed that the CNN approach to document image representation exceeds the power of hand-crafted alternatives.

\section*{Acknowledgements}
This work was supported by NSERC Discovery and Engage grants (held by K.G.D.), and an NSERC USRA (awarded to A.W.H.).  The authors thank Palomino Systems for helpful discussions.  The authors gratefully acknowledge the support of NVIDIA Corporation with the donation of a Tesla K40 GPU used for this research.

{\small
\bibliography{bibref_definitions_short,references}
\bibliographystyle{ieee}
}

\end{document}